\documentclass[times,twocolumn,final,authoryear]{elsarticle}
\usepackage{prletters}
\usepackage{framed,multirow}
\usepackage{amssymb}
\usepackage{latexsym}
\usepackage{graphicx}
\usepackage{amsmath}
\usepackage{adjustbox}
\usepackage{subfig}
\usepackage{cases}
\usepackage{algorithm}
\usepackage{algpseudocode}
\usepackage{pifont}
\usepackage{multirow}
\usepackage{url}
\usepackage{xcolor}
\usepackage{times}
\usepackage{caption}    
\usepackage{multirow}
\usepackage{xcolor}
\usepackage{hyperref}
\usepackage{amsfonts}

\begin{document}
\begin{frontmatter}

\tnotetext[t1]{\textit{This paper is under review to Applied Intelligence journal }}

\title{Learn Class Hierarchy using Convolutional Neural Networks}

\author[1]{Riccardo La Grassa} 
\author[1]{Ignazio Gallo}
\author[1]{Nicola Landro}
\address[1]{University of Insubria, Department of Theoretical and Applied Science, Varese, Italy}

\begin{abstract}
A large amount of research on Convolutional Neural Networks has focused on flat Classification in the multi-class domain. 
In the real world, many problems are naturally expressed as problems of hierarchical classification, in which the classes to be predicted are organized in a hierarchy of classes. 
In this paper, we propose a new architecture for hierarchical classification of images, introducing a stack of deep linear layers with cross-entropy loss functions and center loss combined. 
The proposed architecture can extend any neural network model and simultaneously optimizes loss functions to discover local hierarchical class relationships and a loss function to discover global information from the whole class hierarchy while penalizing class hierarchy violations.
We experimentally show that our hierarchical classifier presents advantages to the traditional classification approaches finding application in computer vision tasks.
\end{abstract}
\end{frontmatter}

\section{Introduction}
In recent years researchers have become increasingly interested in the multi-label  and hierarchical learning approaches, finding many application to several domain, including classification~\cite{wehrmann2018hierarchical,cesa2006incremental}, image annotation~\cite{dimitrovski2011hierarchical}, bioinformatics~\cite{valentini2009true,yan2019zhejiang} \cite{chen2018deep} \cite{abacha2019vqa}.
Nowadays, machine learning is commonly used to resolve complex problem into pattern recognition where an object is classified assigning a label in according with the model's rule used.
However, classes are not always disjoint from others and objects within them can be related to others as a hierarchical structure~\cite{silla2011survey}.
Human beings perceive the world with different types of granularity and can translate information from coarse-grained to fine-grained and on the contrary, perceiving different levels of abstraction of the information acquired~\cite{hobbs1990granularity,mccalla1992granularity}.
This concept is reflected in the taxonomy of the multi-label general approaches under the idea of structured output prediction~\cite{su2015multilabel}.

In terms of neural models, the main difference between the prediction of structured output and flat multi-label classification lies in the level of neurons that contains the label prediction. 
In fact, in the presence of a structured output, the information is based on a different level of abstraction, while with the multi-label flat approach it is based on a single level.

Hierarchical multi-label classification (HMC) is a variant of the classification task where instances may belong to multiple classes at the same time and classes are organized in a hierarchy.
In HMC approaches a relationship among classes and can be formalized by a tree or directed acyclic graph (DAG).
Our approach to HMC exploits the annotation hierarchy by building a single neural network that can simultaneously predict all categorization of an input source exploiting multiple layers of a neural model. 
For example, considering the class label prediction for an image containing a tiger, the proposed system can simultaneously predict that a "tiger" has been found but at the same time the same object is also a "feline" and a "mammal".

In literature exists two main approaches to HMC problem, known as local and global~\cite{costa2007comparing,xu2019survey,silla2011survey}.
In the global approach, the output of the final layer predicts the test instance in which only one classifier sees information globally without having local information.
In the local approach, there is a set of trained classifiers that follows a top-down strategy, in particular, the training process is independently for each base classifier. 

Different local approaches have been proposed in the literature, like Local classifier per Node (LCN) \cite{valentini2009true}, Local classifier per parent node (LCPN), Local classifier per level (LCL) \cite{cerri2011hierarchical}.
LCN strategy trains a local classifier for each node of a graph providing a local decision to make predictions.
LCPN uses a multi-class classifier for each internal class to recognize classes from its sub-classes and LCL methods train a multi-class classifier per hierarchical level.
In contrast with local (LCN, LCL, LCPN) and global approaches, we use a single trained model and a single back-propagation error with many different layers fully connected, responsible to synchronize with a concept linked to a given hierarchical structure.

A recent work~\cite{wehrmann2018hierarchical} describes a novel method to solve HMC problem, that preserves local and global information simultaneously to discover the local hierarchical relationship among classes.
Unlike this work, our architecture exploits recent neural network potentialities and facilitates the multi-class prediction for each deep layers to capture local context following the hierarchical structure of the information. 
In our approach, we have a cascade of fully connected linear layers each one with softmax plus cross-entropy, where the output of a layer $l-1$ is the input of layer $l$; instead, in~\cite{wehrmann2018hierarchical} the model has ReLu activation functions on two different layers fully connected with softmax and binary cross-entropy per block.
Another difference with~\cite{wehrmann2018hierarchical} is that the input of each layer $l$ fuse with the input, instead, in our approach the input per layer is the output of the previous layer.
The last difference is that our model uses local classification as final prediction in according to hierarchical multi-label classification task, instead of in HMCN-F the final layer is used as flat layer plus another layer that uses jointly local and global output information to obtain the final prediction.


Our work can be summarized in the following key contributions:
\begin{itemize}
    \item We propose a new hierarchical deep loss approach (HDL) as an extension of convolutional neural networks to assign hierarchical multi-labels to images. Our extension can be adapted to a generic Convolutional Neural Network as final step.
    \item To prove the effectiveness of our hierarchical classification approach we conduct empirical studies on three different datasets. First, we created \textit{Animals\_Taxonomy8} dataset based on real animal images from Flickr on three groups of taxonomy (Class, Family, Species) with their relative label annotations. Second, we used a well-known biomedical dataset (\textit{VQA-Med 2019}) contains radiology real images on different levels of hierarchy and third, we created \textit{Geometry\_shapes\_annotations} that contains thousands of shapes images on three depth hierarchy levels. Further, all datasets have a different number of instances (2.8k,8k,40k) useful to prove the robustness of our approach.  
    \end{itemize}

\section{The Proposed Approach}
\begin{figure*}
    \centerline{\includegraphics[width=1.0\textwidth]{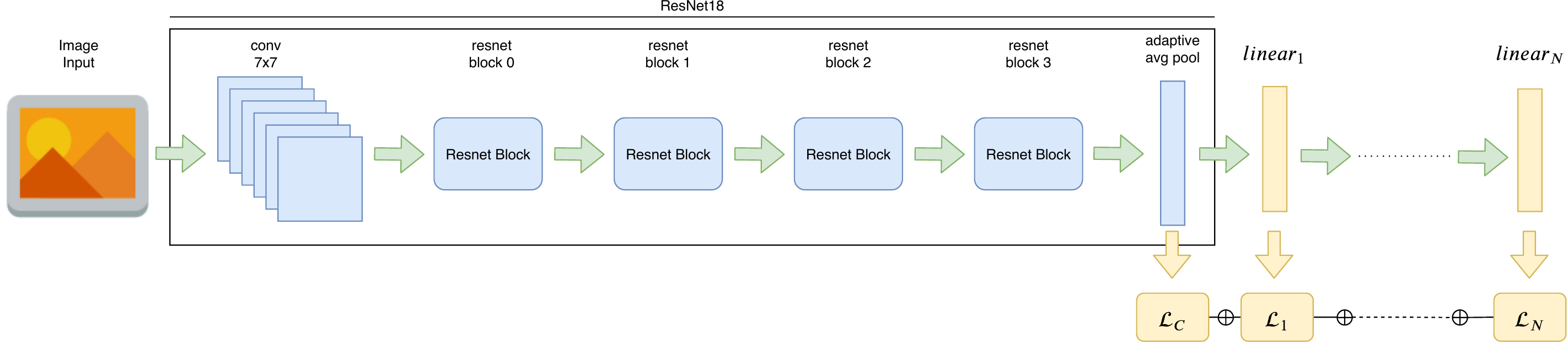}}
    \caption{The proposed architecture of HDL} 
    \label{fig:net}
\end{figure*}

As mentioned above, our solution is an architectural extension that can be adapted to a generic neural network.
In this paper, we used a standard Convolutional Neural Network, the ResNet18, as a base model to which we added our solution to solve a hierarchical images classification problem.
As graphically represented in Figure~\ref{fig:net}, what we do is to extend the output layer with some fully connected layers equal to the number of layers available in the classes hierarchy tree of the problem to be solved, and to associate a loss function to each of these new layers added.
In practice, we construct a mapping between the layers of a class hierarchy and the new layers of the neural network ($linear_1,\dots,linear_N$ in Figure~\ref{fig:net}) so that the network can learn to discriminate between all class labels belonging to a given layer of the hierarchy.
To minimize the intra-class variance and at the same time to keep the features among different classes separated we compute the Center Loss \cite{wen2016discriminative} on each training mini-batch and update all class centers after each training epoch.
More formally we compute the center loss $\mathcal{L}_{C}$ as follow:
\begin{equation}
\mathcal{L}_{C} = \sum_{i=1}^{m}{\lVert x_{i} - c_{y_{i}} \rVert}_{2}^{2} \label{centers_loss}
\end{equation}
where $c_{y_i}\in \mathbb{R}^{d}$ denotes the center for the class $y{_i}$ in the features space of the deep model.
In our experiments, we chose a Resnet-18 as a general model and apply Center Loss after the adaptive pooling layer.
Finally, let $l_{1}$ be the linear layer at first level with dimension equal to the number of classes at first level of hierarchy, more formally:
\begin{equation}
l_{G}^{1} = \phi(W_{G}^{1} x + b_{G}^{1})
\label{linear_layer}
\end{equation}
where $W_{G}^{1} \in \mathbb{R}^{|l_{G}^{1}| \times |d|}$ , $b_{G}^{1} \in \mathbb{R}^{|l_{G}^{1}| \times 1}$ is the bias vector with $\phi$ linear activation function and $d$ be the number of features.
Then, we add a linear layer $l$ for each hierarchical level in a generic dataset and we perform the cross-entropy loss to maximize the inter-class variance.
Precisely, we apply softmax function from logits of layer $l_{G}^{l}$ and use cross-entropy loss as Eq.\ref{cross-entropy-with-softmax_loss}
\begin{equation}
  \mathcal{L}_{l_{G}^{l}} = - \sum_{i=1}^{m} log \frac{e^{W_{y_{i}}^{T}x_i+b_{yi}}}{\sum_{j=1}^{n} e^{W_{j}^{T}x_i+b_j}}
  \label{cross-entropy-with-softmax_loss}
\end{equation}

Where, $l$ is the layer l-th, m and n are the mini-batch size and number of classes respectively, $x_i \in \mathbb{R}^d$ denotes the \textit{i}th deep feature, belonging to the $y_i$th class and b is the bias. 

Finally, our total loss is:
\begin{eqnarray}
    \mathcal{L} = \lambda_{0} \cdot \mathcal{L}_{C} + \lambda_1 \cdot \mathcal{L}_{1} + \dots + \lambda_N \cdot \mathcal{L}_{N}
    \label{total_loss}
\end{eqnarray}
Where $\lambda_{0,1,...,N}=1$,
$\mathcal{L}_{C} $ it the centers loss value and $\mathcal{L}_{0,1,..,N} $ is the cross-entropy loss value of the layer $\{1,...,N\}$.
The general formulation with $N$ layer is defined as Eq.\ref{general_total_loss}
\begin{eqnarray}
    \mathcal{L} =  \lambda_{0} \cdot \mathcal{L}_{C} + \sum_{l=1}^{N} {\lambda_l \cdot \mathcal{L}_{l}}
    \label{general_total_loss}
\end{eqnarray}


\section{Datasets}
To evaluate the proposed method, we created our own datasets as there is no a standard benchmarked dataset on hierarchical multi-label images classification, available in the literature.

\begin{figure}
    \centering
    \subfloat[t1, sagittal, skull and contents]{
         \includegraphics[width=30mm,height=30mm]{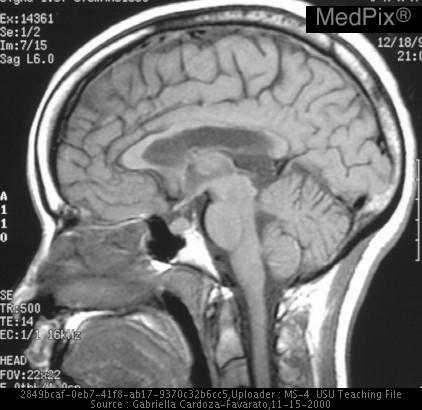}
    }
    \subfloat[xr-plain film, lateral, lung-mediastinum-pleura]{
         \includegraphics[width=30mm,height=30mm]{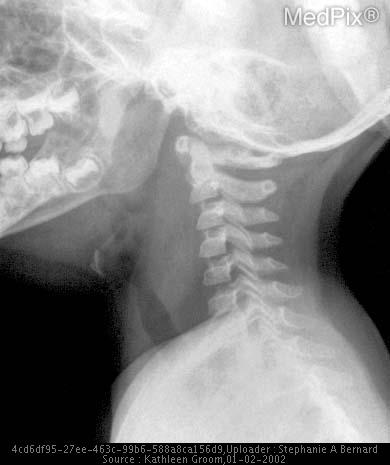}
    }\\
    \subfloat[xr-plain film, ap, musculoskeletal]{
         \includegraphics[width=30mm,height=30mm]{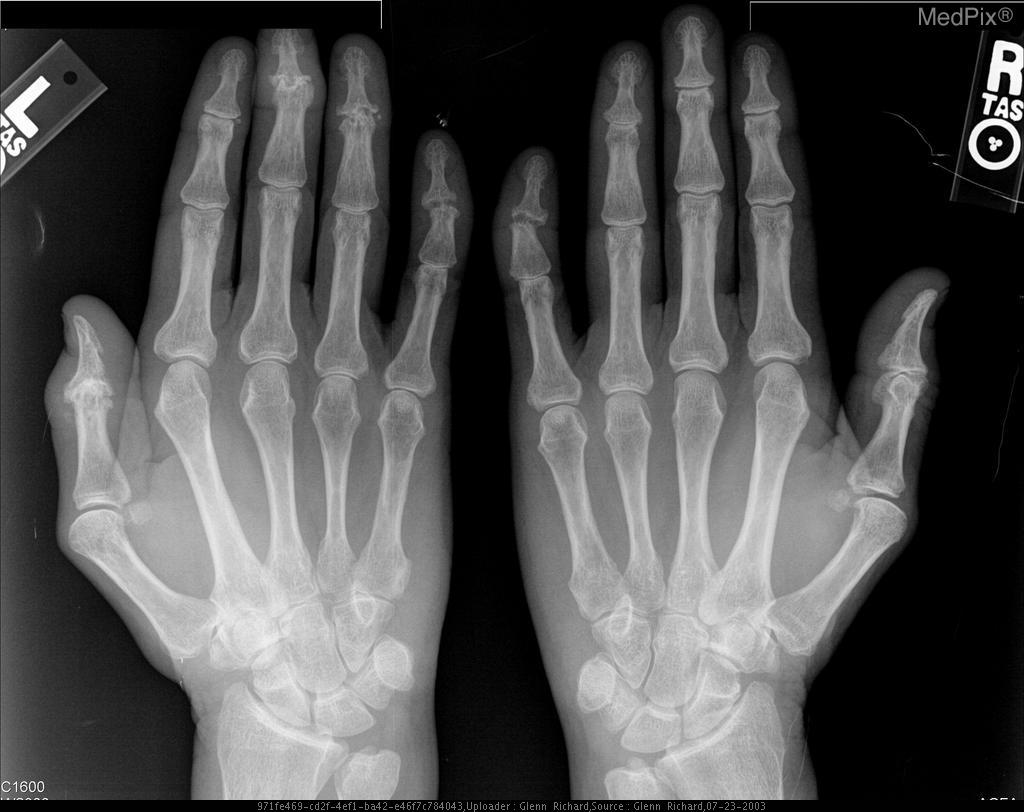}
    }
    \subfloat[ct w/contrast (iv), axial, skull and contents]{
         \includegraphics[width=30mm,height=30mm]{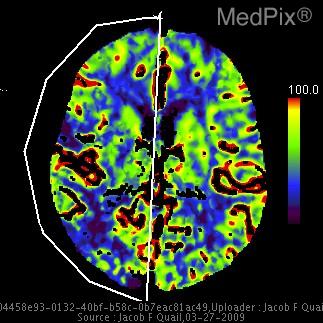}
    }
    \caption{Four images extracted from VQA-Med 2019 \cite{abacha2019vqa} dataset (synpic371, synpic10103, synpic16486, synpic48315). The labels separated by comma belong to the sub-categories of the three main classes following the order: Modality, Plain, Organ.} 
    \label{fig:biomedical_dataset}
\end{figure}

The medical Visual Question Answering task (VQA-Med 2019)~\cite{abacha2019vqa} is focused on radiology images (example in Fig. \ref{fig:biomedical_dataset}) grouped in four main classes: Modality, Plane, Organ system, Abnormality. The original challenge is to classify an image from a question linked to it, indeed for each image in the training we have a paired question. 
Our focus is on the hierarchical multi-label classification of images, therefore, we will exclude our experiment from text classification task.
We use all train size and use the validation set as a test set (because the test set is not labelled with all labels), respectively 2816/340 objects.
In total, we consider three levels of hierarchy (Modality Class, Plane Class, Organ Class) with their relative different type of concepts.
These classes have a size of 44, 15, 10 respectively per classes.
In these experiments, our goal is to prove experimentally the effectiveness and robustness of our model to discriminate different concepts also in the case we have a few examples per classes in the train.

\begin{figure}
    \centering
    \subfloat[Green square with orange border]{
         \includegraphics[width=0.1\textwidth]{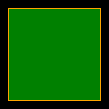}
    }
    \hspace{0.01\textwidth}
    \subfloat[Gray square with red border]{
        \includegraphics[width=0.1\textwidth]{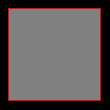}
    }
     \subfloat[Green triangle with red border]{
         \includegraphics[width=0.1\textwidth]{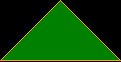}
    }
    \hspace{0.01\textwidth}
    \subfloat[Gray triangle with red border]{
        \includegraphics[width=0.1\textwidth]{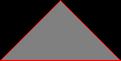}
    }
    \caption{Some sample of geometric dataset.} 
    \label{fig:geometricdataset}
\end{figure}

We have created a synthetic geometric shapes dataset which contains 2 different shapes (Triangle, Square, some image sample into fig. \ref{fig:geometricdataset}) at the first level of our hierarchy. Each shape has 6 different full colours and other 6 different colours for out-fill, the last two represents the second and third level of the hierarchy. The possible configuration is 72 so, we generate train/test with 20000 and 6000 objects respectively.
The dimension of the images is 128x128x3. In these experiments, we want to answer the question "Which kind of shape is this? What is the fill colour? and the out fill colour?.

\begin{figure*}
    \centerline{\includegraphics[width=1.\textwidth]{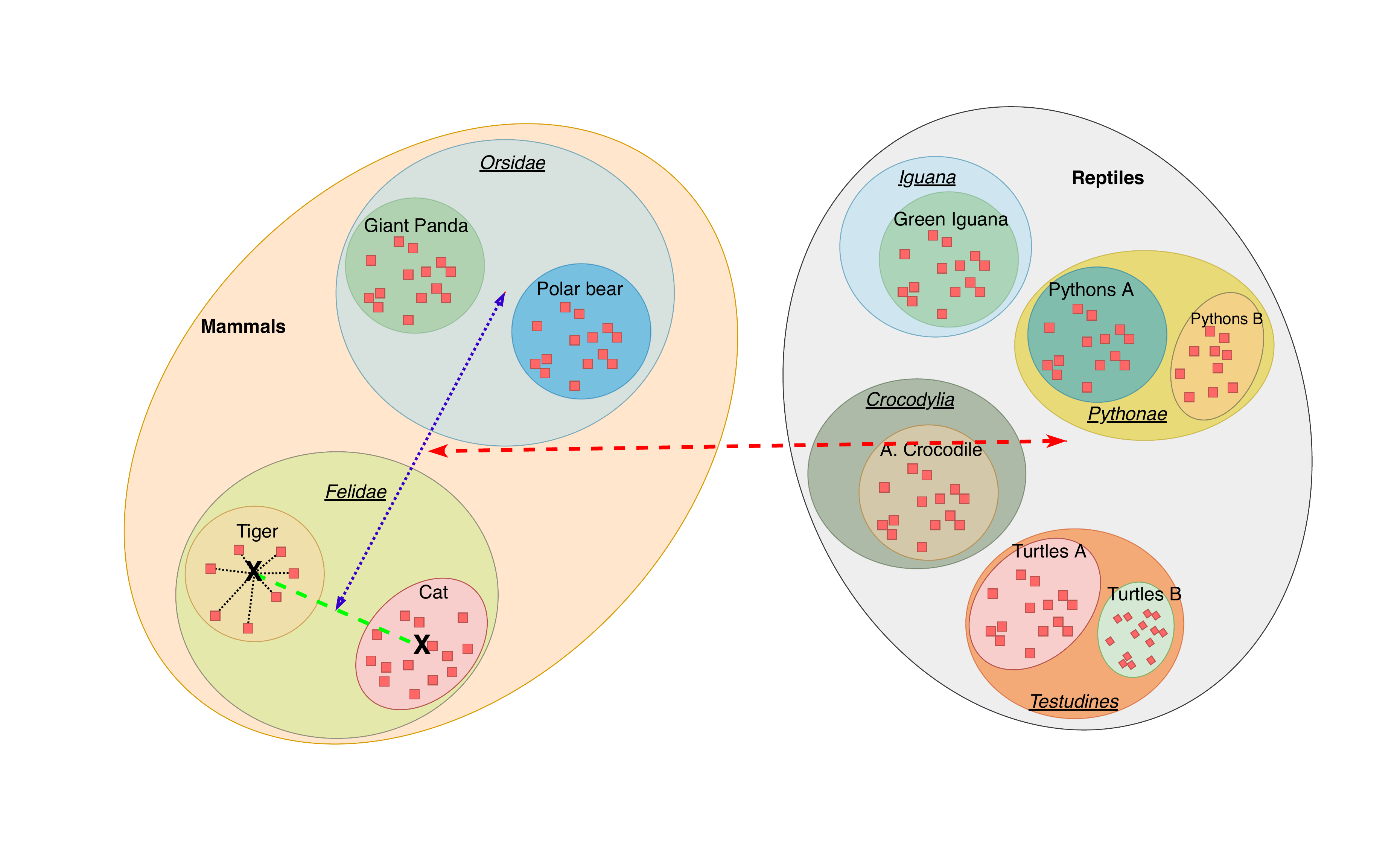}}
    \caption{Hierarchy of categories used in \textit{Animals\_Taxonomy8}} \label{fig:Hierarchy_images}
\end{figure*}

\begin{figure}
    \centering
    \subfloat[Felis catus]{
         \includegraphics[height=1.1cm]{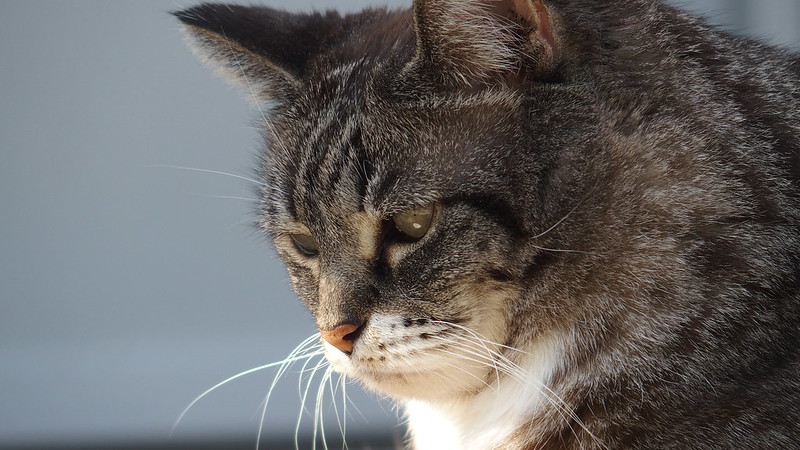}
    }
    \hspace{0.05cm}
    \subfloat[Malaysia tiger]{
         \includegraphics[height=1.1cm]{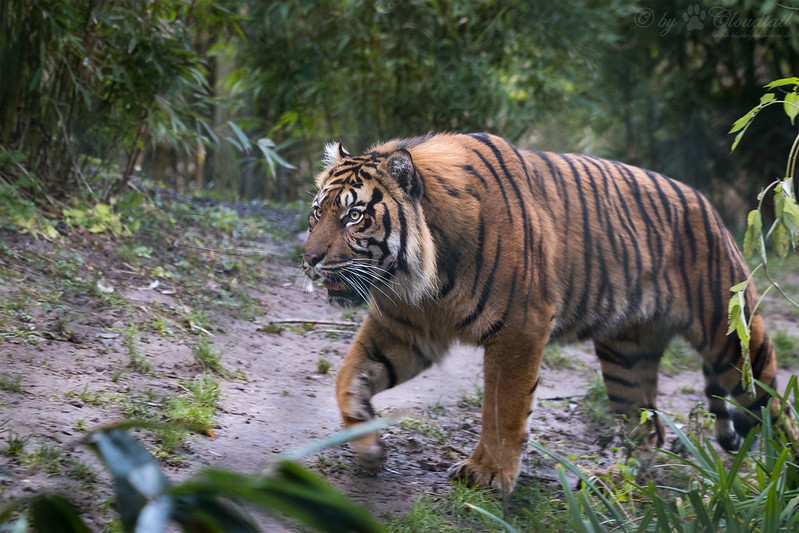}
    }
    \hspace{0.05cm}
    \subfloat[Ursus maritimus]{
         \includegraphics[height=1.1cm]{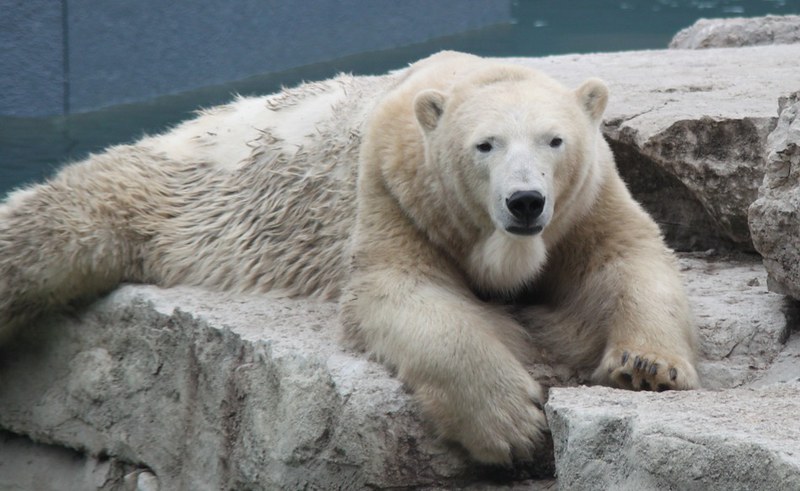}
    }
    \hspace{0.05cm}
    \subfloat[Ailuropoda malanoleuca]{
         \includegraphics[height=1.1cm]{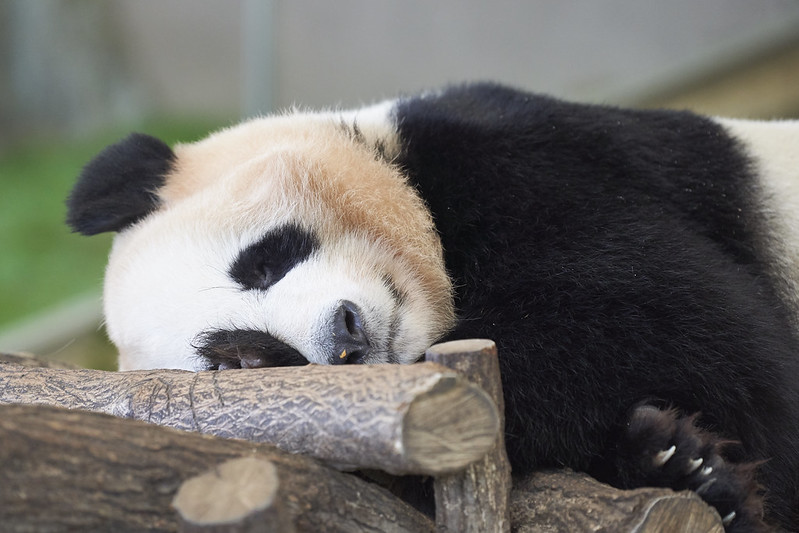}
    }
    
    \subfloat[Python molurus]{
         \includegraphics[height=1.1cm]{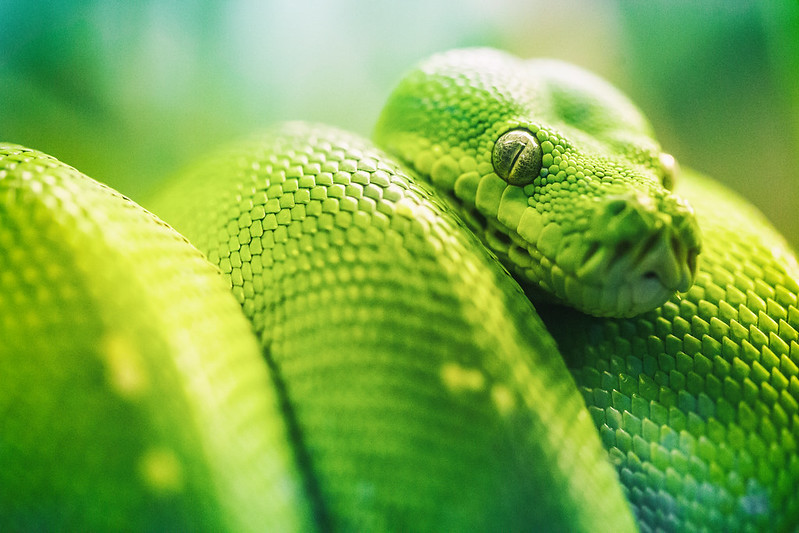}
    }
    \hspace{0.05cm}
    \subfloat[Trachemys scripta]{
         \includegraphics[height=1.1cm]{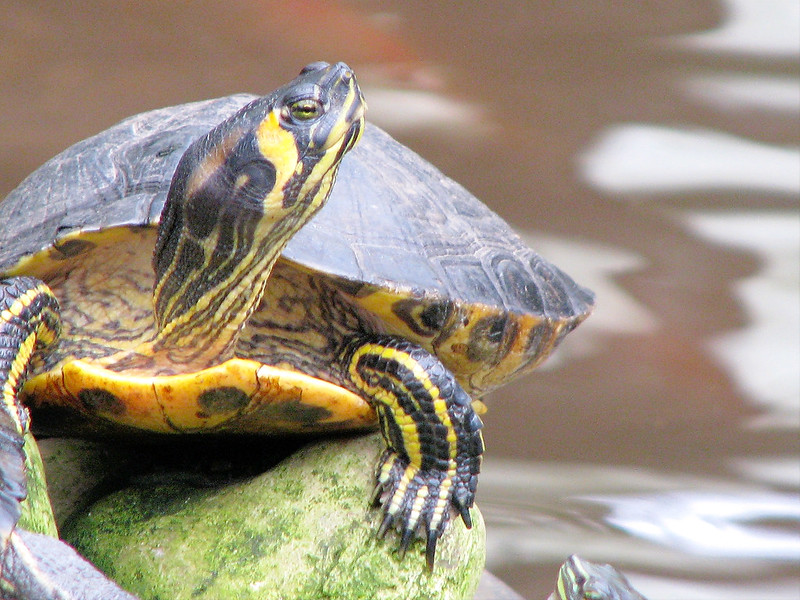}
    }
    \hspace{0.05cm}
    \subfloat[Green iguana]{
         \includegraphics[height=1.1cm]{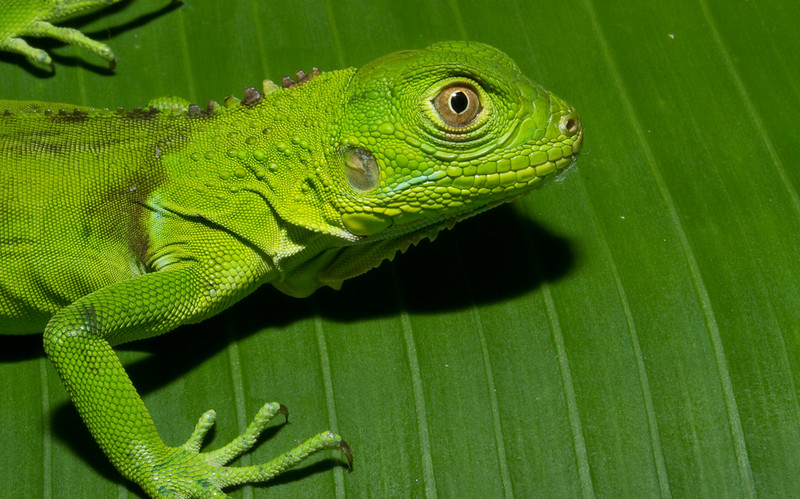}
    }
    \hspace{0.05cm}
    \subfloat[Crocodylus niloticus]{
         \includegraphics[height=1.1cm]{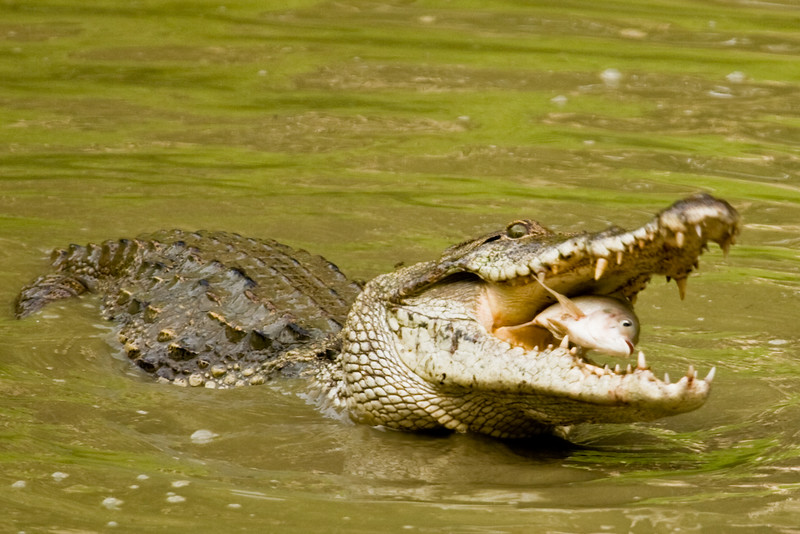}
    }
    \caption{Example of images extracted from \textit{Animals\_Taxonomy8}} 
    \label{fig:animals_dataset}
\end{figure}

\begin{figure}
    \centerline{\includegraphics[width=0.7\columnwidth]{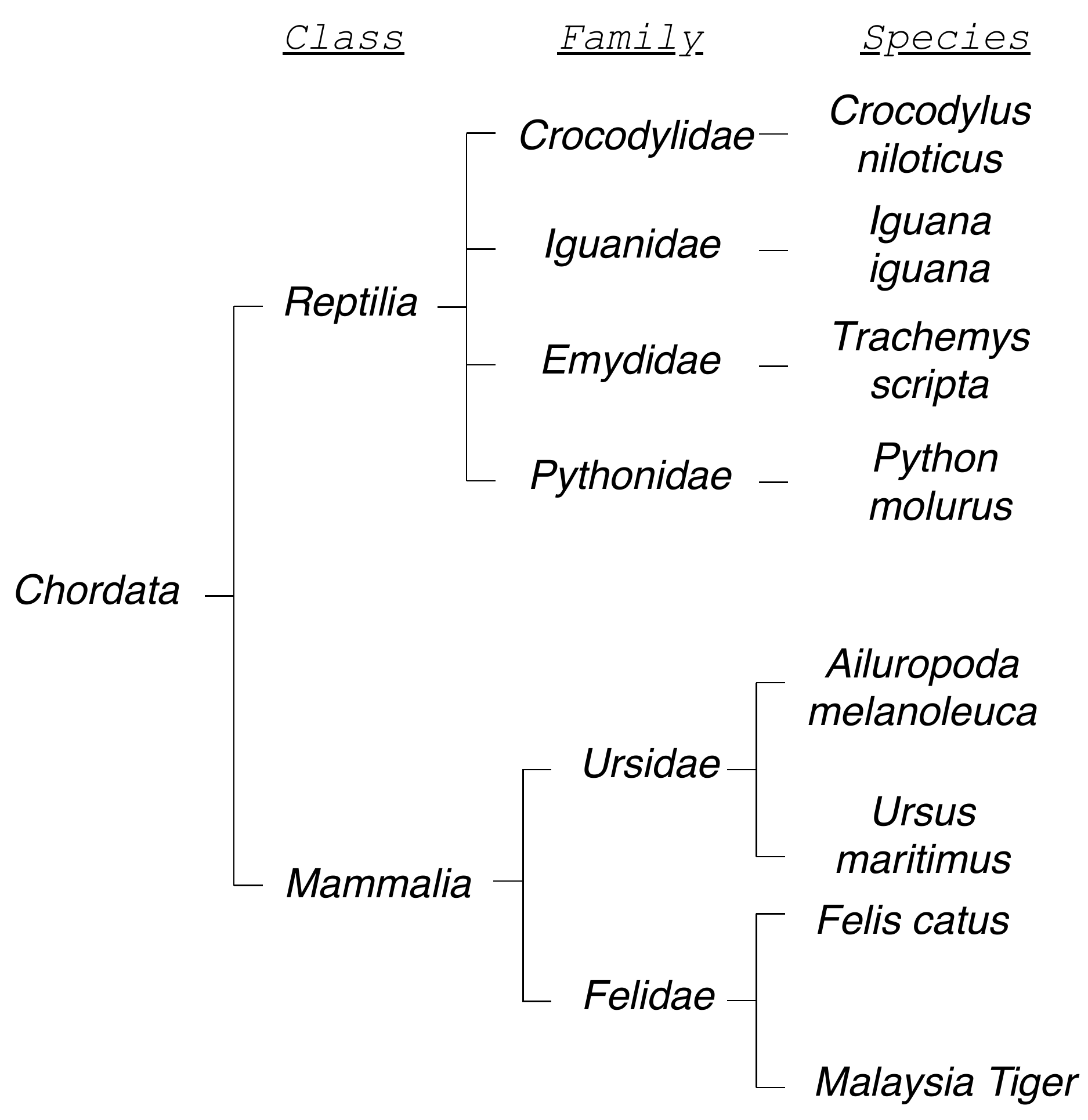}}
    \caption{\textit{Animals\_Taxonomy8} dataset classification hierarchy.} 
    \label{fig:animal_class_schema}
\end{figure}

This data set is created from Flickr animals images, the hierarchy represents a small taxonomy with class, family and species as in Fig. \ref{fig:animal_class_schema}.
The selected \textbf{class} is \textit{mammalia} and \textit{reptilia}. 
The second level of hierarchy is the \textbf{family}, in particular \textit{felidae} and \textit{ursidae} for mammalia and \textit{crocodyle}, \textit{iguanidae}, \textit{emydidae} and \textit{pythonidae}  for reptilia.
The last hierarchy level represent the \textbf{species}( example of images in fig. \ref{fig:animals_dataset}) as \textit{malaysia tiger}, \textit{felis catus} known as cat, \textit{ailuropoda melanoleuca} known as giant panda,
\textit{ursus maritimus} known as polar bear, \textit{python molurus} known as green python, \textit{trachemys scripta} as small turtle, \textit{iguana iguana} and \textit{crocodylus niloticus} well known as nilus crocodile.
A whole representation of the dataset is in Fig. \ref{fig:Hierarchy_images}.

\section{Experiments}
To evaluate the proposed method we develop four empirical studies.
\begin{enumerate}
    \item In the first one, we use a well-known dataset (VQA-Med 2019) to test our approach with biomedical real images, also in the case we have few data available.
    \item In the second we test the capability of abstraction of our approach on a synthetic dataset created in the context we have thousand of instances available.
    \item In the third, we extract hierarchical structure on the real-images dataset contains images of three types of animal taxonomy levels (Class-Family-Species) and prove the robustness of our HDL in the case which images are hard to recognize and they contain noise.
    \item In the four experiment, we compare our HDL with a ResNet18 proving the effectiveness of our approach.
\end{enumerate}{}

\begin{figure}
\centerline{\includegraphics[width=\columnwidth]{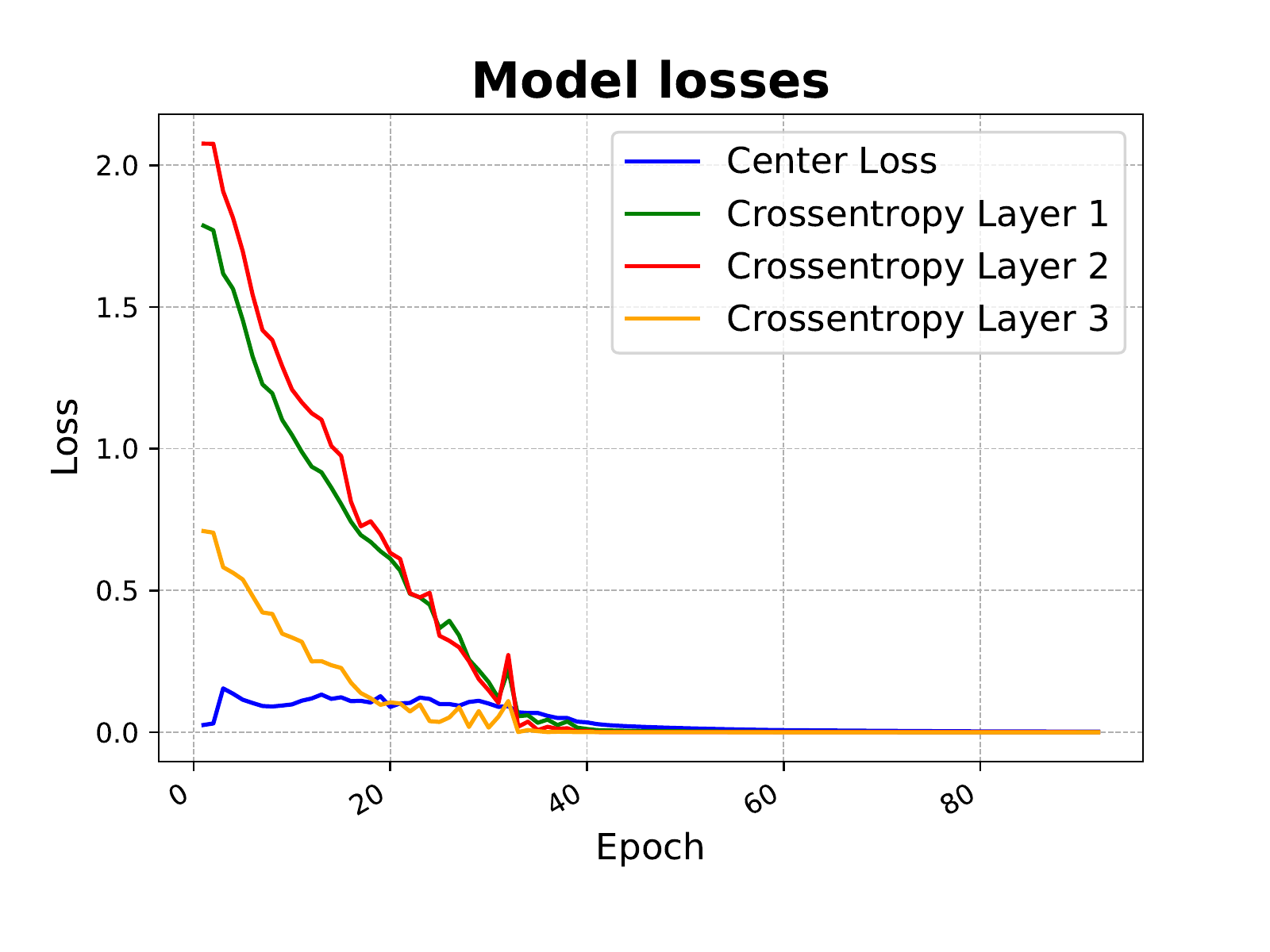}}
\caption{Training losses \textit{Animals\_Taxonomy8} with $lr=0.01$} \label{fig:animal_losses_1}
\end{figure}

\begin{figure}
\centerline{\includegraphics[width=\columnwidth]{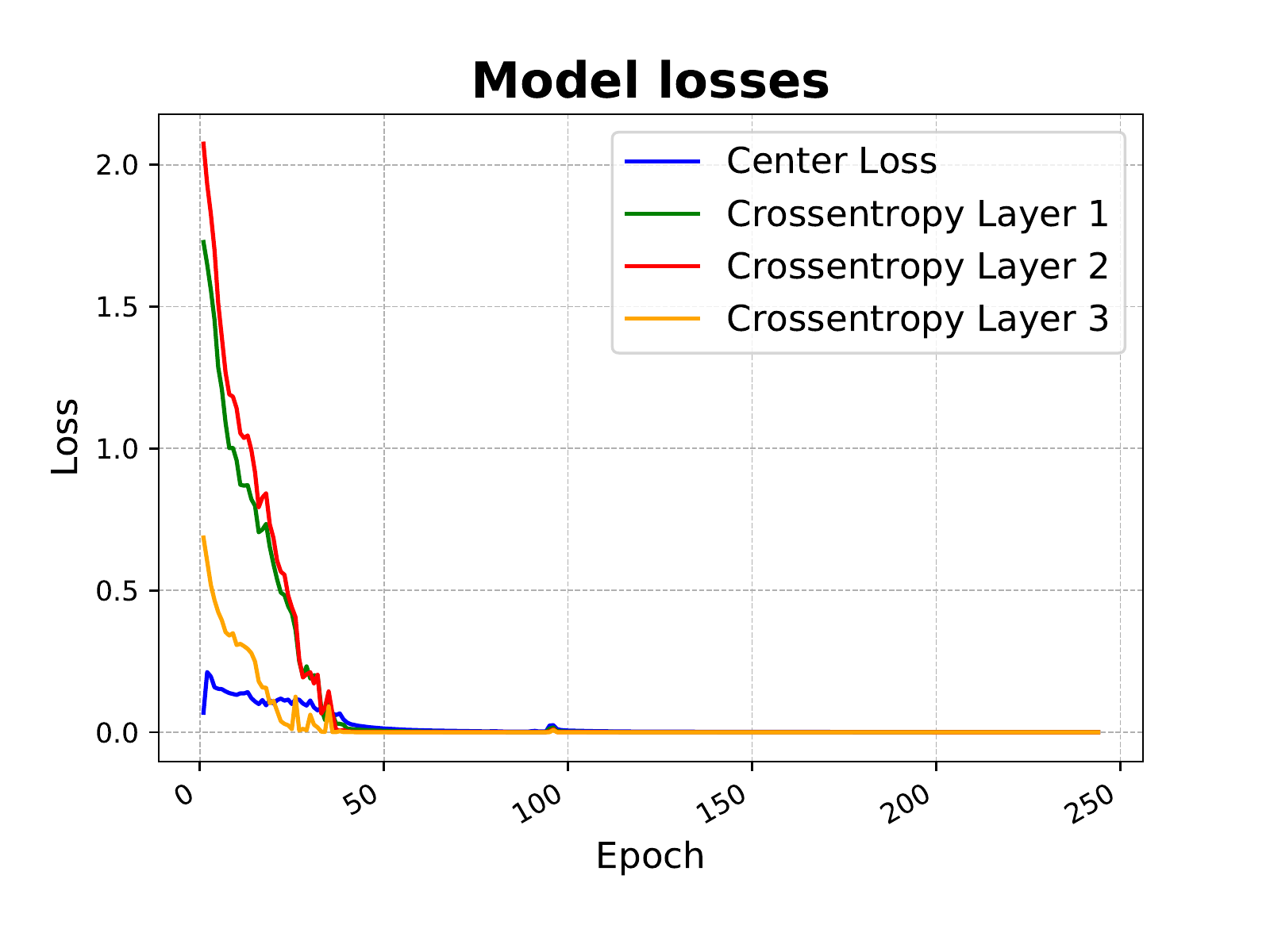}}
\caption{Training losses \textit{Animals\_Taxonomy8} with $lr=0.005$.
We emphasize the different descent of losses. This is due to the number of concepts to distinguish from each layer. Each line represents the loss for each layer. For this dataset, we design our model with a shape of 6,8,2 to distinguish Family, Species and Classes respectively. As we show also in \ref{table1_results}, the line yellow that represents linear layer 3 with 2 concepts (Mammals or Reptiles) has more descent power, indicating that our model quickly learns a few concepts rather than many as red line or green.} \label{fig:animal_losses_5}
\end{figure}

\textbf{First experiment}
In this experiment, we test our model in the case we have few instances and with a high complexity of images. Our hypothesis is that the performance in terms of accuracy in a layer is higher when the number of different concepts to distinguish is inferior to a layer with many concepts to recognize. As we show in \ref{table1_results} at the first row (VQA-Med 2019), we have accuracies of 38.05, 74.04, 66.66 for the size of layers 44, 15, 10 respectively. We can observe that the accuracy of the first layer is lower of 1.94 times than the second layer and to 1.75 times than the third layer, this proves that our model offers better scalability when we have few concepts per layer to learn. Similar results can be found in \textit{Animals\_Taxonomy8}, where the higher accuracy of the third layer at the third row of Table \ref{table1_results} than others, is due to the fact we have only two concept (mammals or reptiles) to distinguish than the second layer (8 concepts) Figs. \ref{fig:animal_losses_1} and \ref{fig:animal_losses_5}.

\textbf{Second experiment}
In a second experiment, we use a synthetic dataset with simple geometric shapes and several instances 7.10 times greater than VQA-Med 2019.
Our intuition is that attribute more samples per classes can improve the training of our model and subsequently, to obtain better performance in terms of accuracy than the first experiments. To prove this conjecture, we train with 20K instances our HDL and test it with 6k instances. The results in Table. \ref{table1_results} at the second row per tables, confirm our expectations. The higher number of instances jointly with the simplicity of images allows the model to reach high accuracy starting from the first ten epochs. Furthermore, we conduct three different runs with a learning rate of 0.005, 0.001, 0.01 using batch-size of 64.

\textbf{Third experiment}
In these experiments, we test our model using more instances than the first experiment and with images of animal (\textit{Animals\_Taxonomy8}) that contains noise.
In particular, our model offers good performance also in the case the images are not simple as in the second experiments and when they contain noise or offers little comprehensibility, indeed many images are not clear, like for example a snake completely hidden by forest or a bar sign with a panda logo. However, as we show in \ref{table1_results}, the accuracy of the third layer, responsible to recognize mammals or reptile is very high. We conclude that considering the poor understanding of images, noise and hard images to recognize, experimental results prove the robustness of our model.

\begin{figure}
\centerline{\includegraphics[width=\columnwidth]{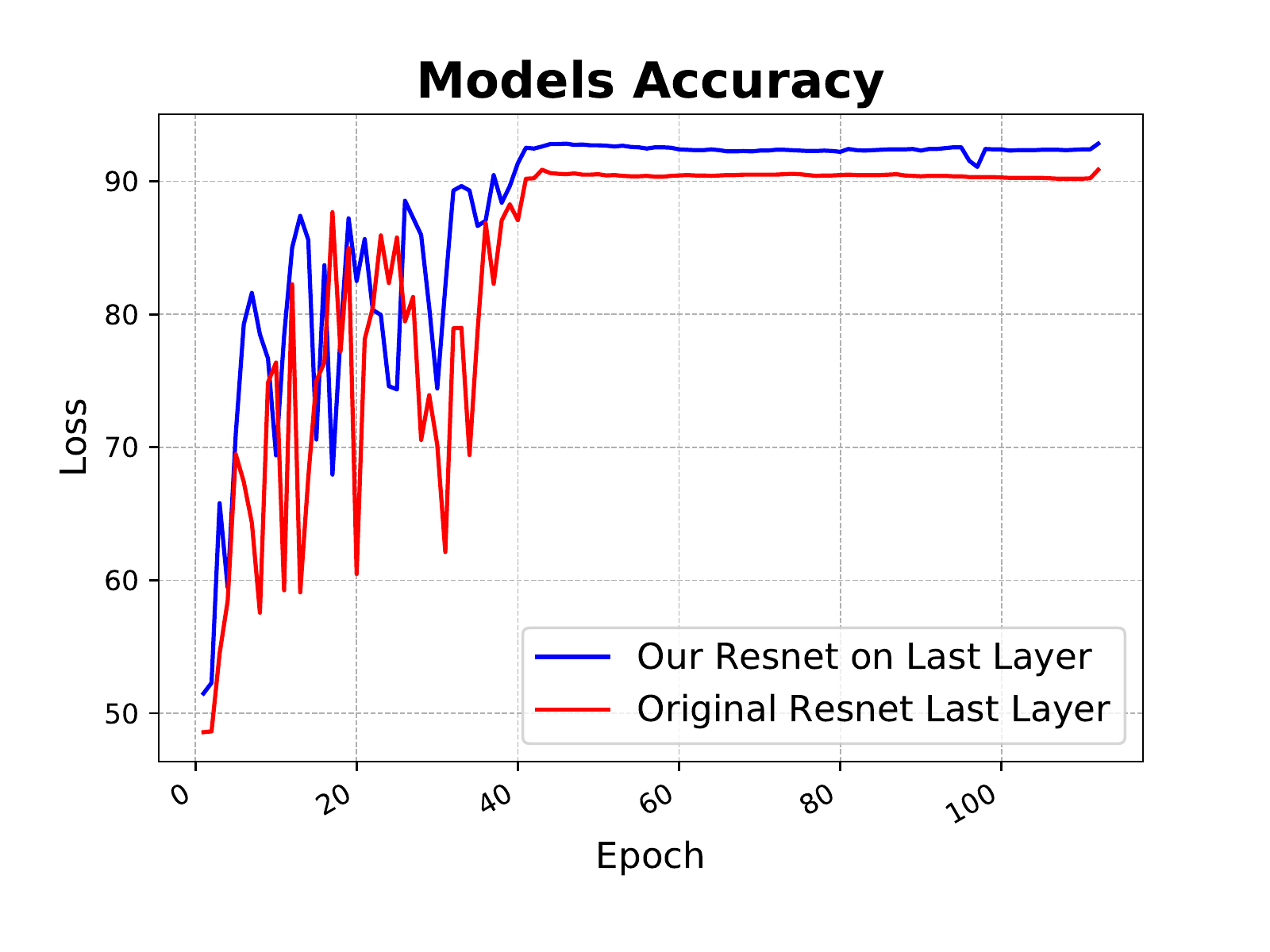}}
\caption{HDL vs original Resnet18 with $lr=0.005$} \label{fig:our_vs_standard}
\end{figure}

\begin{figure}
\centerline{\includegraphics[width=\columnwidth]{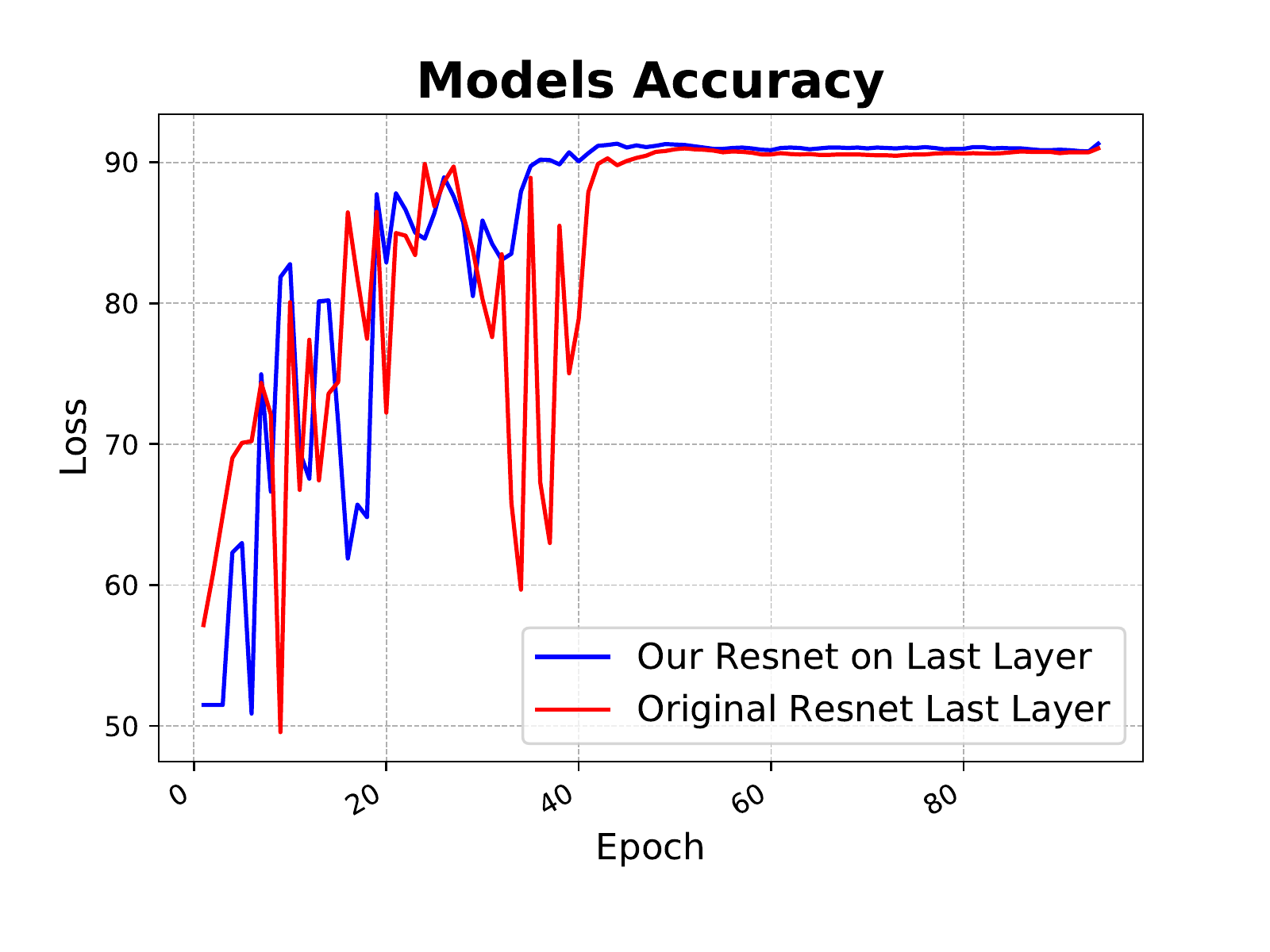}}
\caption{HDL vs original Resnet18 with $lr=0.01$} \label{fig:our_vs_standard_f2}
\end{figure}

\textbf{Fourth experiment}
HDL is designed to maximize the learning capacity and to extract the hierarchical structure from the labelled data. Our intuition is that our model, lead to different losses at any level, with the power to reduce intra-variance and to maximize inter-variance, can obtain better accuracy than a classical convolutional neural network.
To prove this, we conduct six different experiments using a classic ResNet18 and our HDL on \textit{Animals\_Taxonomy8} using two learning rate and a batch size of 64. The results in Fig. \ref{table2_comparison_results} and \ref{fig:our_vs_standard},\ref{fig:our_vs_standard_f2} clearly confirm our expectations. In all cases, the accuracy is higher than a classical ResNet18, this experiment proves the effectiveness of our proposed model.

\subsection{Experiments settings}
We build our hierarchical multi-label classifier model as an extension on a Resnet-18, but is it possible to apply to any Convolutional Neural networks.
We implement our extension in Python using Pytorch framework. Fig.\ref{fig:net} shows the architecture used for experiments. The size of the input images is re-scaled to 64x64x3 for Geometry dataset and 256x256x3 for \textit{VQA-Med 2019} and \textit{Animals\_Taxonomy8} datasets.
We do not apply any preprocessing of images as data augmentation, rotation or normalization. The kernel size of the first convolutional layers is 7x7 with a stride of 2 pixels, followed by a normalization of layer and a non-linear layer with ReLu activation function. A max-pooling operation over 3x3 regions and a stride of 2 pixels. Then, we have four blocks of Convolution, with 64, 128, 256, 512 numbers of plans respectively and apply an adaptive average pooling over 1x1 region. Finally, we add three fully connected linear layer, where each layer corresponding to the total number of concepts in our hierarchical dataset. In the forward process, we take the output after the adaptive average pooling and apply Center loss function and for each linear layers we apply softmax function and then cross-entropy loss. The total loss will be the sum of the local loss per layers. Our network was trained with Adam optimizer \cite{kingma2014adam}.
The batch-size used, learning rate, epochs are described jointly with the results for each dataset.

\section{Results and Discussion}
This study is placed in the sub-category of multi-label classification called Structure output learning. In according with experimental results at Tables \ref{table1_results}, \ref{table2_comparison_results}, we achieved good results on three different datasets finding the way to exploit the dependency among classes and make accurate predictions, reducing the misclassification than a classic ResNet18.
The main reason we have created these datasets is to prove our proposal in the field of computer vision and with more than 2 levels of depth, indeed CIFAR100 contains only two levels of depth (Super Class, Classes) and other datasets with many depths find applicability only in text classification or in bioinformatics, where the inputs are not images.

\begin{table}
\begin{center}
{\caption{Accuracies comparison using three different datasets on different learning rate. We use a batch size of 32 on VQA-MED and 64 on the other datasets.}\label{table1_results}}
\begin{tabular}{lccc}
\hline
\rule{0pt}{8pt}
&\multicolumn{3}{c}{lr=0.005}\\
\cline{1-4}
\rule{0pt}{10pt}
\quad Datasets&1l&2l&3l\\
\hline
\\[-6pt]
\quad \textit{VQA-Med} &38.05\% &74.04\% &66.66\% \\
\quad \textit{Shapes}  &100\% &100\% &100\%\\
\quad \textit{Animals\_Taxonomy8} &71.98\% &69.07\% &92.82\%\\
\hline
\rule{0pt}{8pt}
&\multicolumn{3}{c}{lr=0.001}\\
\hline
\\[-6pt]
\quad \textit{VQA-Med}  &35.98\% &70.20\%   &67.84\%\\
\quad \textit{Shapes}   &100\% &100\%   &100\%\\
\quad \textit{Animals\_Taxonomy8}  &72\% &69.12\%   &92.89\%\\
\hline
\rule{0pt}{8pt}
&\multicolumn{3}{c}{lr=0.01}\\
\hline
\\[-6pt]
\quad \textit{VQA-Med} &34.8\% &71.97\% &69.61\%\\
\quad \textit{Shapes}  &100\% &100\% &100\%\\
\quad \textit{Animals\_Taxonomy8} &69.2\% &66.53\% &91.32\%\\
\end{tabular}
\end{center}
\end{table}

\begin{table*}
\begin{center}
{\caption{Accuracies comparison of our model with a original ResNet on different layers. The batch size used for these experiments is 64.}\label{table2_comparison_results}}
\begin{tabular}{lccc|c}
\hline
\rule{0pt}{8pt}
&\multicolumn{2}{c}{lr=0.005}\\
\cline{0-4}
\rule{0pt}{10pt}
\quad \textbf{Our Model}&1l&2l&3l&ResNet18
\\
\hline
\\[-6pt]
                                  &\textbf{71.98\%} &- &- &71.19\% \\
\quad \textit{Animals\_Taxonomy8} &- &\textbf{69.07\%} &- &68.58\% \\
                                  &- &- &\textbf{92.82\%} &90.86\% \\
\hline

\rule{0pt}{8pt}
&\multicolumn{2}{c}{lr=0.01}\\
\cline{0-4}
\rule{0pt}{10pt}
\quad \textbf{Our Model}&1l&2l&3l&ResNet18
\\
\hline
\\[-6pt]
                                           &\textbf{69.2\%}&- &- &68.34\%  \\
\quad \textit{Animals\_Taxonomy8}          &- &\textbf{66.53\%} &- &65.36\% \\
                                           &- &- &\textbf{91.32\%} &90.98\% \\
\hline

\end{tabular}
\end{center}
\end{table*}

\section{Conclusion}
In literature, multi-label classification is an important field in machine learning and it is strongly related to many real-world applications for example, in biomedical images annotation, document categorization and whatever problem which the instances inside the classes are not disjoint but they keep a hierarchical structure. In this work, we have conducted four empirical studies on different datasets to prove by experimental results the effectiveness and robustness of our proposed model, that can be applied as an extension to any Convolutional Neural Network.

\bibliographystyle{model2-names}
\bibliography{main}

\begin{thebibliography}{16}
\expandafter\ifx\csname natexlab\endcsname\relax\def\natexlab#1{#1}\fi
\providecommand{\url}[1]{\texttt{#1}}
\providecommand{\href}[2]{#2}
\providecommand{\path}[1]{#1}
\providecommand{\DOIprefix}{doi:}
\providecommand{\ArXivprefix}{arXiv:}
\providecommand{\URLprefix}{URL: }
\providecommand{\Pubmedprefix}{pmid:}
\providecommand{\doi}[1]{\href{http://dx.doi.org/#1}{\path{#1}}}
\providecommand{\Pubmed}[1]{\href{pmid:#1}{\path{#1}}}
\providecommand{\bibinfo}[2]{#2}
\ifx\xfnm\relax \def\xfnm[#1]{\unskip,\space#1}\fi
\bibitem[{Abacha et~al.(2019)Abacha, Hasan, Datla, Liu, Demner-Fushman and
  M{\"u}ller}]{abacha2019vqa}
\bibinfo{author}{Abacha, A.B.}, \bibinfo{author}{Hasan, S.A.},
  \bibinfo{author}{Datla, V.V.}, \bibinfo{author}{Liu, J.},
  \bibinfo{author}{Demner-Fushman, D.}, \bibinfo{author}{M{\"u}ller, H.},
  \bibinfo{year}{2019}.
\newblock \bibinfo{title}{Vqa-med: Overview of the medical visual question
  answering task at imageclef 2019}, in: \bibinfo{booktitle}{CLEF2019 Working
  Notes. CEUR Workshop Proceedings}, pp. \bibinfo{pages}{09--12}.
\bibitem[{Cerri et~al.(2011)Cerri, Barros and
  de~Carvalho}]{cerri2011hierarchical}
\bibinfo{author}{Cerri, R.}, \bibinfo{author}{Barros, R.C.},
  \bibinfo{author}{de~Carvalho, A.C.}, \bibinfo{year}{2011}.
\newblock \bibinfo{title}{Hierarchical multi-label classification for protein
  function prediction: A local approach based on neural networks}, in:
  \bibinfo{booktitle}{2011 11th International Conference on Intelligent Systems
  Design and Applications}, \bibinfo{organization}{IEEE}. pp.
  \bibinfo{pages}{337--343}.
\bibitem[{Cesa-Bianchi et~al.(2006)Cesa-Bianchi, Gentile and
  Zaniboni}]{cesa2006incremental}
\bibinfo{author}{Cesa-Bianchi, N.}, \bibinfo{author}{Gentile, C.},
  \bibinfo{author}{Zaniboni, L.}, \bibinfo{year}{2006}.
\newblock \bibinfo{title}{Incremental algorithms for hierarchical
  classification}.
\newblock \bibinfo{journal}{Journal of Machine Learning Research}
  \bibinfo{volume}{7}, \bibinfo{pages}{31--54}.
\bibitem[{Chen et~al.(2018)Chen, Miao, Xu, Hager and Harrison}]{chen2018deep}
\bibinfo{author}{Chen, H.}, \bibinfo{author}{Miao, S.}, \bibinfo{author}{Xu,
  D.}, \bibinfo{author}{Hager, G.D.}, \bibinfo{author}{Harrison, A.P.},
  \bibinfo{year}{2018}.
\newblock \bibinfo{title}{Deep hierarchical multi-label classification of chest
  x-ray images} .
\bibitem[{Costa et~al.(2007)Costa, Lorena, Carvalho, Freitas and
  Holden}]{costa2007comparing}
\bibinfo{author}{Costa, E.P.}, \bibinfo{author}{Lorena, A.C.},
  \bibinfo{author}{Carvalho, A.C.}, \bibinfo{author}{Freitas, A.A.},
  \bibinfo{author}{Holden, N.}, \bibinfo{year}{2007}.
\newblock \bibinfo{title}{Comparing several approaches for hierarchical
  classification of proteins with decision trees}, in:
  \bibinfo{booktitle}{Brazilian Symposium on Bioinformatics},
  \bibinfo{organization}{Springer}. pp. \bibinfo{pages}{126--137}.
\bibitem[{Dimitrovski et~al.(2011)Dimitrovski, Kocev, Loskovska and
  D{\v{z}}eroski}]{dimitrovski2011hierarchical}
\bibinfo{author}{Dimitrovski, I.}, \bibinfo{author}{Kocev, D.},
  \bibinfo{author}{Loskovska, S.}, \bibinfo{author}{D{\v{z}}eroski, S.},
  \bibinfo{year}{2011}.
\newblock \bibinfo{title}{Hierarchical annotation of medical images}.
\newblock \bibinfo{journal}{Pattern Recognition} \bibinfo{volume}{44},
  \bibinfo{pages}{2436--2449}.
\bibitem[{Hobbs(1990)}]{hobbs1990granularity}
\bibinfo{author}{Hobbs, J.R.}, \bibinfo{year}{1990}.
\newblock \bibinfo{title}{Granularity}, in: \bibinfo{booktitle}{Readings in
  qualitative reasoning about physical systems}. \bibinfo{publisher}{Elsevier},
  pp. \bibinfo{pages}{542--545}.
\bibitem[{Kingma and Ba(2014)}]{kingma2014adam}
\bibinfo{author}{Kingma, D.P.}, \bibinfo{author}{Ba, J.}, \bibinfo{year}{2014}.
\newblock \bibinfo{title}{Adam: A method for stochastic optimization}.
\newblock \bibinfo{journal}{arXiv preprint arXiv:1412.6980} .
\bibitem[{McCalla et~al.(1992)McCalla, Greer, Barrie and
  Pospisil}]{mccalla1992granularity}
\bibinfo{author}{McCalla, G.}, \bibinfo{author}{Greer, J.},
  \bibinfo{author}{Barrie, B.}, \bibinfo{author}{Pospisil, P.},
  \bibinfo{year}{1992}.
\newblock \bibinfo{title}{Granularity hierarchies}.
\newblock \bibinfo{journal}{Computers \& Mathematics with Applications}
  \bibinfo{volume}{23}, \bibinfo{pages}{363--375}.
\bibitem[{Silla and Freitas(2011)}]{silla2011survey}
\bibinfo{author}{Silla, C.N.}, \bibinfo{author}{Freitas, A.A.},
  \bibinfo{year}{2011}.
\newblock \bibinfo{title}{A survey of hierarchical classification across
  different application domains}.
\newblock \bibinfo{journal}{Data Mining and Knowledge Discovery}
  \bibinfo{volume}{22}, \bibinfo{pages}{31--72}.
\bibitem[{Su et~al.(2015)}]{su2015multilabel}
\bibinfo{author}{Su, H.}, et~al., \bibinfo{year}{2015}.
\newblock \bibinfo{title}{Multilabel classification through structured output
  learning-methods and applications} .
\bibitem[{Valentini(2009)}]{valentini2009true}
\bibinfo{author}{Valentini, G.}, \bibinfo{year}{2009}.
\newblock \bibinfo{title}{True path rule hierarchical ensembles}, in:
  \bibinfo{booktitle}{International Workshop on Multiple Classifier Systems},
  \bibinfo{organization}{Springer}. pp. \bibinfo{pages}{232--241}.
\bibitem[{Wehrmann et~al.(2018)Wehrmann, Cerri and
  Barros}]{wehrmann2018hierarchical}
\bibinfo{author}{Wehrmann, J.}, \bibinfo{author}{Cerri, R.},
  \bibinfo{author}{Barros, R.}, \bibinfo{year}{2018}.
\newblock \bibinfo{title}{Hierarchical multi-label classification networks},
  in: \bibinfo{booktitle}{International Conference on Machine Learning}, pp.
  \bibinfo{pages}{5225--5234}.
\bibitem[{Wen et~al.(2016)Wen, Zhang, Li and Qiao}]{wen2016discriminative}
\bibinfo{author}{Wen, Y.}, \bibinfo{author}{Zhang, K.}, \bibinfo{author}{Li,
  Z.}, \bibinfo{author}{Qiao, Y.}, \bibinfo{year}{2016}.
\newblock \bibinfo{title}{A discriminative feature learning approach for deep
  face recognition}, in: \bibinfo{booktitle}{European conference on computer
  vision}, \bibinfo{organization}{Springer}. pp. \bibinfo{pages}{499--515}.
\bibitem[{Xu et~al.(2019)Xu, Shi, Tsang, Ong, Gong and Shen}]{xu2019survey}
\bibinfo{author}{Xu, D.}, \bibinfo{author}{Shi, Y.}, \bibinfo{author}{Tsang,
  I.W.}, \bibinfo{author}{Ong, Y.S.}, \bibinfo{author}{Gong, C.},
  \bibinfo{author}{Shen, X.}, \bibinfo{year}{2019}.
\newblock \bibinfo{title}{A survey on multi-output learning}.
\newblock \bibinfo{journal}{arXiv preprint arXiv:1901.00248} .
\bibitem[{Yan et~al.(2019)Yan, Li, Xie, Xiao and Gu}]{yan2019zhejiang}
\bibinfo{author}{Yan, X.}, \bibinfo{author}{Li, L.}, \bibinfo{author}{Xie, C.},
  \bibinfo{author}{Xiao, J.}, \bibinfo{author}{Gu, L.}, \bibinfo{year}{2019}.
\newblock \bibinfo{title}{Zhejiang university at imageclef 2019 visual question
  answering in the medical domain}.
\newblock \bibinfo{journal}{Working Notes of CLEF} .

\end{thebibliography}
\end{document}